%% file: acl_latex.tex
\pdfoutput=1
\documentclass[11pt]{article}

\usepackage{acl}

\usepackage{times}
\usepackage{latexsym}

\usepackage{soul}
\usepackage{amsmath}
\usepackage{booktabs}
\usepackage{bm}
\usepackage{cleveref}
\usepackage{graphicx}
\usepackage{tabularx}
\usepackage{soul}
\usepackage{xcolor}
\usepackage{todonotes}
\usepackage{multirow}
\usepackage{xpatch}
\usepackage{blindtext}
\usepackage{fdsymbol}

\usepackage{microtype}
\usepackage{hyperref}
\usepackage{adjustbox}
\usepackage{textcomp}
\usepackage{xparse}
\usepackage{enumitem}

\usepackage[T1]{fontenc}
\usepackage[utf8]{inputenc}
\usepackage[arabic,farsi,russian,english]{babel}

\usepackage{microtype}

\usepackage{xparse}
\usepackage[noorphans,vskip=0.5ex]{quoting}

\input{contents/commands}

\iffalse

\NewDocumentCommand{\heng}
{ mO{} }{\textcolor{red}{\textsuperscript{\textit{Heng}}\textsf{\textbf{\small[#1]}}}}
\NewDocumentCommand{\steeve}
{ mO{} }{\textcolor{gold}{\textsuperscript{\textit{Steeve}}\textsf{\textbf{\small[#1]}}}}
\else
\newcommand{\heng}[1]{}
\newcommand{\xueqing}[1]{}
\newcommand{\yi}[1]{}
\newcommand{\steeve}[1]{}
\fi
\title{\modelshort~: Improving Cross-lingual Fact-checking with \\Cross-lingual Retrieval}

\author{Kung-Hsiang Huang ~~~ ChengXiang Zhai ~~~ Heng Ji\\
  Department of Computer Science, University of Illinois Urbana-Champaign \\
  \texttt{\{khhuang3, czhai, hengji\}@illinois.edu} \\}

\begin{document}
\maketitle

\input{contents/00_abstract}
\input{contents/01_introduction}
\input{contents/03_task}
\input{contents/04_method}
\input{contents/05_experiment}
\input{contents/06_results}

\input{contents/02_related_work}
\input{contents/07_conclusion}
\input{contents/08_ethics}
\section*{Acknowledgement}
This research is based upon work supported by U.S. DARPA SemaFor Program No. HR001120C0123. The views and conclusions contained herein are those of the authors and should not be interpreted as necessarily representing the official policies, either expressed or implied, of DARPA, or the U.S. Government. The U.S. Government is authorized to reproduce and distribute reprints for governmental purposes notwithstanding any copyright annotation therein.
\bibliography{anthology,custom}
\bibliographystyle{acl_natbib}

\end{document}

%% file: contents/commands.tex
\newcommand{\xfact}[1]{\textsc{X-Fact}}
\newcommand{\modelshort}[1]{\textsc{Concrete}}
\newcommand{\xict}[1]{\textsc{X-ICT}}
\newcommand{\mdpr}[1]{mDPR}

\definecolor{c2}{RGB}{218,0,0}

\definecolor{lightblue}{RGB}{212, 235, 255}
\definecolor{lightorange}{RGB}{255, 204, 168}
\definecolor{lightyellow}{RGB}{255, 255, 168}
\definecolor{gold}{rgb}{0.83, 0.69, 0.22}
\sethlcolor{lightblue}

%% file: contents/00_abstract.tex
\begin{abstract}

Fact-checking has gained increasing attention due to the widespread of falsified information. Most fact-checking approaches focus on claims made in English only due to the data scarcity issue in other languages. The lack of fact-checking datasets in low-resource languages calls for an effective cross-lingual transfer technique for fact-checking. Additionally, trustworthy information in different languages can be complementary and helpful in verifying facts. To this end, we present the first fact-checking framework augmented with cross-lingual retrieval that aggregates evidence retrieved from multiple languages through a cross-lingual retriever. Given the absence of cross-lingual information retrieval datasets with claim-like queries, we train the retriever with our proposed \textit{Cross-lingual Inverse Cloze Task} (\xict~), a self-supervised algorithm that creates training instances by translating the title of a passage. The goal for \xict~ is to learn cross-lingual retrieval in which the model learns to identify the passage corresponding to a given translated title. On the \xfact~ dataset, our approach achieves 2.23\% absolute F1 improvement in the zero-shot cross-lingual setup over prior systems. The source code and data are publicly available at \url{https://github.com/khuangaf/CONCRETE}. 
\end{abstract}

%% file: contents/01_introduction.tex
\section{Introduction}

Fact-checking is an important task that assesses the veracity of a claim. This task has gained increasing attention due to the widespread mis- and dis-information that has a significant socioeconomic impact on our society \cite{scheufele2019science, pate2019impact, fung-etal-2021-infosurgeon, 10.1162/tacl_a_00454, wu-etal-2022-cross, fung-etal-2022-the, huang2022faking}. While fact-checking is mainly conducted manually, especially in the journalism industry, with more than hundred of millions of social media posts and millions of blog posts published per day \cite{hoang2018location, firstsiteguide_2022}, manual fact-checking is no longer feasible. Hence, we are urgently in need of reliable automated fact-checking approaches.

\begin{figure*}[t]
    \centering
    \includegraphics[width=0.9\linewidth]{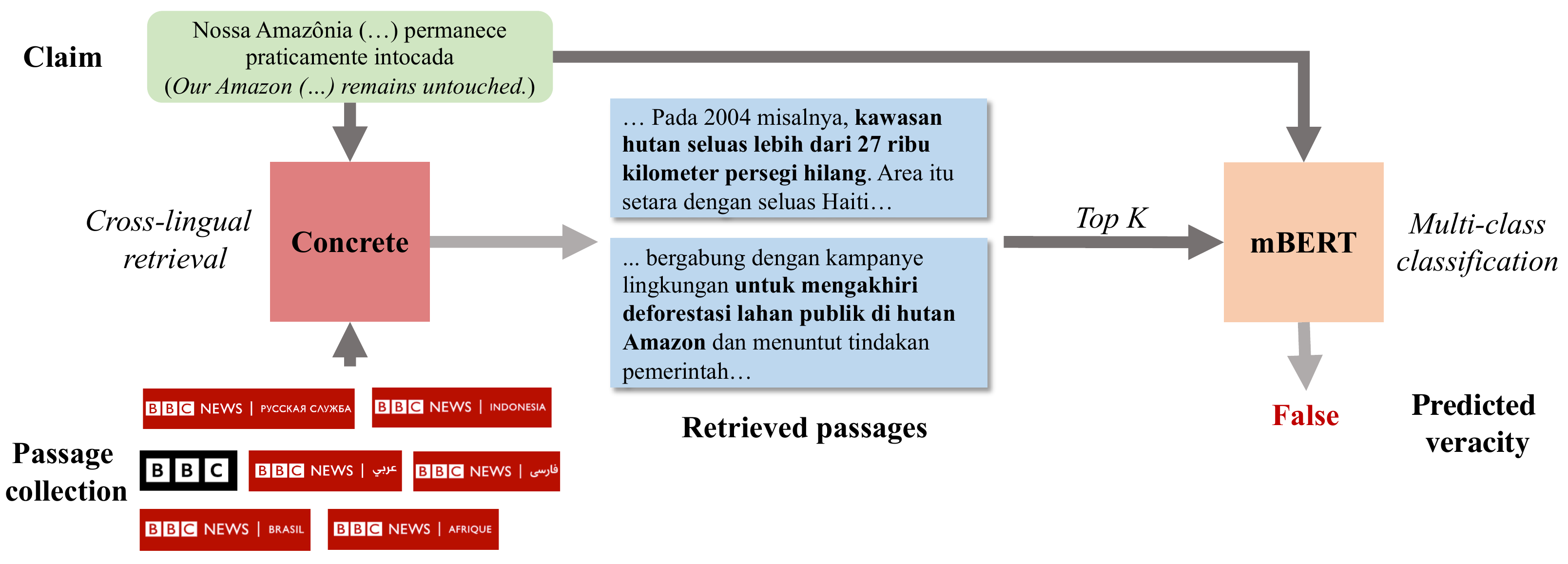}
    \vspace{-2mm}
    \caption{An overview of the proposed framework. Given a claim in arbitrary language, a cross-lingual retriever, \modelshort~ retrieves relevant passages in \textit{any languages}. The top-$k$ relevant passages and the claim are then passed to our multilingual reader, mBERT, to predict the veracity of the claim. }
    \vspace{-5mm}
    \label{fig:framework_overview}
\end{figure*}
Most existing work develops fact-checking approaches on English-only corpus \cite{wang-2017-liar, thorne-etal-2018-fever, augenstein-etal-2019-multifc, wadden-etal-2020-fact}. One reason for this is the scarcity of fact-checking websites in other languages for constructing large enough non-English fact-checking datasets. Therefore, it is even more challenging to build fact-checkers for low-resource languages. A solution is to leverage high-resource languages with zero-shot cross-lingual transfer, where the model is trained on source languages of richer resources and directly tested on target languages of lower resources. Note that the sets of languages in the training set and the test sets are disjoint. With this technique, ground-truth labels for claims in low-resource languages are no longer required.

Few studies in the fact-checking literature have explored the cross-lingual setup. One line of work attempts to match an input claim with claims in other languages that have been verified \cite{kazemi-etal-2021-claim, kazemi2022matching}. However, this approach fails when the claims have not been fact-checked in any language. Another line of work builds classifiers with multilingual language models. For example, \citet{gupta-srikumar-2021-x} utilize Google Search to obtain snippets that are relevant to the input claim and train a model based on mBERT \cite{devlin-etal-2019-bert} in a cross-lingual setting. Although Google Search excels at retrieving relevant information from a given claim, it disregards the trustworthiness of the information being retrieved. When a claim is erroneous, the search results often contradict each other, which likely leads to incorrect predictions, as shown later in \Cref{subsec:qualitative_analysis}.

Motivated by these challenges, we propose \modelshort~, a \textbf{\underline{C}}laim-\textbf{\underline{o}}rie\textbf{\underline{n}}ted \textbf{\underline{C}}oss-lingual \textbf{\underline{Ret}}ri\textbf{\underline{e}}ver that retrieves evidence from a trustworthy multilingual passage collection for fact-checking. This approach can handle region-specific claims as long as relevant evidence is presented in the passage collection, which is much more accessible compared to similar claims in other languages. In addition, since it does not rely on a black-box retrieval system, our approach provides the flexibility to include only trustworthy information in the passage collection. One major challenge for training such a retriever is the lack of multilingual information retrieval (IR) dataset with claim-like queries. To this end, we propose a self-supervised cross-lingual learning algorithm, Cross-lingual Inverse Cloze Task (\xict~), to learn the retriever based on pseudo-feedback. To mimic cross-lingual retrieval, we construct a pseudo-query for a given passage by translating its title into a randomly selected language. The objective for \xict~ is to identify the passage corresponding to a given translated title among all candidate passages. Since the title of news articles can often be regarded as a claim, this approach mitigates the domain discrepancy issue. As shown in \Cref{fig:framework_overview}, our framework first performs cross-lingual retrieval to obtain evidence relevant to the input claim. Then, a multilingual reader takes in the retrieved evidence and the input claim to classify the veracity of the claim.

Our contributions can be summarized as follows:

\begin{itemize}[noitemsep,nolistsep]
    \item To the best of our knowledge, we present the first fact-checking framework augmented with cross-lingual retrieval that achieves state-of-the-art cross-lingual transfer performance on the \xfact~ fact-checking task.
  \item We propose \modelshort~, a cross-lingual retriever with a bi-encoder architecture learned through a proposed self-supervised learning algorithm.
  
  \item Our experiments reveal that the distance between the input claim and the retrieved passages is strongly correlated with the performance.
  \item We collected a multilingual passage collection composed of reliable news articles in seven different languages. We have demonstrated that this corpus is effective for retrieval-augmented fact-checking.
  
\end{itemize}

%% file: contents/03_task.tex
\section{Task Definitions}

The input is a claim $c$ in an arbitrary language and the corresponding metadata, such as claimer and claim date. Based on $c$, the retriever component of our model retrieves relevant passages $p$ from a multilingual passage collection $P$ where $p$ can be in any language. Then, the reader component takes in the claim $c$, the corresponding metadata, and relevant passages $p$ to predict the veracity of $c$. Note that the use of passage collection $P$ to aid in fact-checking is a modeling choice, as no grounded evidence is available.

We aim to build a passage collection that does not contain erroneous information. Hence, we construct $P$ by crawling 49,000 articles published in 7 languages\footnote{We consider Arabic, Russian, Indonesian, Persian, French, and Portuguese.} between September 2016 and December 2022 from a trustworthy news media, \url{bbc.com}%\footnote{\url{bbc.com} is considered highly trustworthy according to multiple media rating platforms, such as \url{mediabiasfactcheck.com}.}
, and split each article into passages. Each passage contains at most 100 tokens, following \citet{karpukhin-etal-2020-dense}. This results in a total of 347,557 passages. The collected passage collection has been made publicly available in the link mentioned in the Abstract.

%% file: contents/04_method.tex
\section{Proposed Method}
\label{sec:method}
Our framework is a pipeline consisting of two components: (1) \modelshort~, a claim-oriented cross-lingual retriever that retrieves relevant passages from a multilingual passage collection, and (2) a multilingual reader that determines the veracity of a claim based on the compatibility of the claim and the passages retrieved. \Cref{fig:framework_overview} shows an overview of the proposed method. The following sections describe each component in detail.

\subsection{\modelshort~}
\modelshort~ takes in a claim $c$ in an \textit{arbitrary language} as input and retrieves $k$ relevant passages $p = \{p_0, ... ,p_k\}$ from a multilingual passage collection $\mathcal{P}$, where the retrieved passages can be in \textit{any language}. We adapt \mdpr~ \cite{asai2021cora}, a multilingual retriever, with the proposed self-supervised learning algorithm, Cross-lingual Inverse Cloze Task, for claim-like queries.

\paragraph{\mdpr~} \mdpr~ is a multilingual version of the Dense Passage Retriever (DPR)  \cite{karpukhin-etal-2020-dense}. Similar to DPR, \mdpr~ is a bi-encoder architecture that computes the relevance score between a query and a passage with the inner product of the corresponding representations. In addition to being trained on existing multilingual question answering datasets, \mdpr~ also learns from additional samples mined from Wikipedia and labeled with the proposed answer generator discussed in \citet{asai2021cora}. However, since \mdpr~ was trained on datasets where queries are questions instead of claims, domain mismatch becomes an issue if we directly apply \mdpr~ to our task. 

\paragraph{Cross-lingual Inverse Cloze Task} To address the domain discrepancy problem discussed in the previous paragraph, a naive solution is to fine-tune \mdpr~ on multilingual IR datasets with claim-like queries. Unfortunately, such datasets are not available. Motivated by the Inverse Cloze Task \cite{lee-etal-2019-latent}, which was used to warm start the retriever by tasking a retriever to predict the context given a randomly sampled sentence, we propose Cross-lingual Inverse Cloze Task (\xict~) by extending ICT to a cross-lingual setup. Specifically, we made two major modifications to the original ICT. 

First, instead of randomly sampling a sentence as the pseudo query, we treat the title of each passage as the pseudo query as these titles are often claim-like sentences. The positive passages for a given title are the passages derived from the same article. The biggest advantage of this approach is that there is very little domain mismatch between using claims as queries and using titles as queries. Therefore, the retriever optimized with our proposed \xict~ can be directly applied to claim-oriented downstream tasks without further fine-tuning, indicating that the optimization for the downstream tasks can be more efficient. Second, to mimic cross-lingual retrieval, we use mBART-50 \cite{tang2020multilingual}, a machine translation model, to translate the title into a target language. 

Formally, in \modelshort~, two dense encoders, $E_C(\cdot)$ and $E_P(\cdot)$, are used to represent claims and passages as $d$-dimensional vectors. The similarity between a claim and a passage is defined as the dot product between their dense vectors,
\begin{align}
    sim(c, p_i) = E_C(c)^\top E_P(p_i).
\end{align}
In \xict~, we form positive claim-passage pairs by constructing the translation of the corresponding title $T_{p_i}$ of a given passage $p_i$. i.e. ($T'_{p_i}$, $p_i$) where $T'_{p_i}$ is the original title $T_{p_i}$ translated into a different language using mBART-50. We treat other passages in the same batch as negative samples. The retriever is trained by optimizing the negative log likelihood,

\begin{figure}[t]
    \centering
    \includegraphics[width=0.9\linewidth]{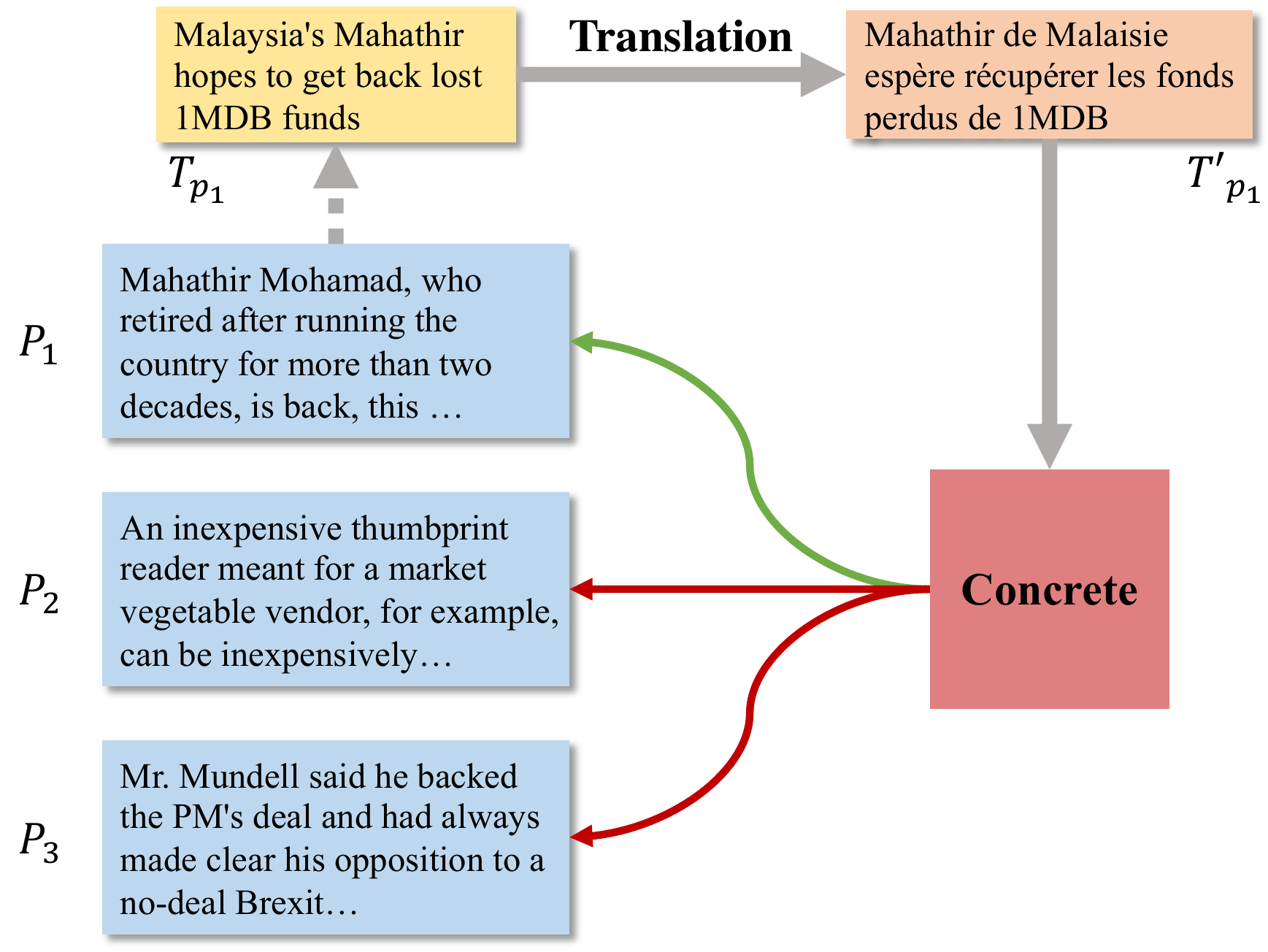}
    \vspace{-2mm}
    \caption{An illustration of \xict~. Given a passage $p_1$, we find its title $T_{p_1}$ and translates it into a different language $T'_{p_1}$ to mimic cross-lingual retrieval. The goal for the cross-lingual retriever, \modelshort~, is to select the correct passage $p_1$ based on the translated title $T'_{p_1}$ among all the passages in the same batch.}
    \vspace{-5mm}
    \label{fig:xict}
\end{figure}

\begin{align}
    P(p_i| T'_{p_i}) &=  \frac{\exp(sim(T'_{p_i}, p_i))}{\sum_{p_j \in \textsc{batch}} \exp(sim(T'_{p_i}, p_j))} \\
    \mathcal{L_{\xict~}} &= - \log \sum_{p_i \in \mathcal{P}} P(p_i | T'_{p_i})
\end{align}
\Cref{fig:xict} demonstrates a graphical illustration.

However, if the translation is performed on every claim, the model would be discouraged from retrieving passages in the same language as the query, which is not a desirable property of the retriever. Therefore, we set an equal probability for a claim to be translated to any language or not being translated (i.e. the probability of not doing translation is $\frac{1}{7}$). We repurposed the passage collection as the training corpus for learning X-ICT.

\subsection{Multilingual Reader}
We use mBERT \cite{devlin-etal-2019-bert}, a multilingual version of BERT, as the encoder for our multilingual reader. The claim $c$ and the corresponding metadata are encoded jointly with a template $T$: [Claim made by \textit{Claimer} on \textit{Claim-Date}, reported in \textit{Language}: {\textit{Claim}}], where \textit{Claimer}, \textit{Claim-Date}, and \textit{Language} are placeholders. The claim template $T$ and each of the $k$\footnote{Empirically, we set $k$ to 5 for best overall performance.} retrieved passages $p_i$ are first encoded independently
\begin{align}
    h_T &= \textrm{mBERT}(T)\texttt{[CLS]}\\
    h_{p_i} &= \textrm{mBERT}(p_i)\texttt{[CLS]}.
\end{align}
The final prediction is then made by feeding the concatenated {\tt[CLS]} embeddings into a multi-layer perceptron $\hat{y} = \textrm{MLP}([h_T; h_{p_0}; ~... ~;h_{p_k}])$. The model is optimized with the cross-entropy loss
\begin{align}
    \mathcal{L} &= \frac{1}{N}\sum_{i=1}^{N}y_i \log \hat{y}_i,
\end{align}
where $y_i$ and $\hat{y}_i$ denote the ground truth label and the predicted label of the $i$-th sample respectively, and $N$ denotes the total number of samples.

%% file: contents/05_experiment.tex
\input{tables/dataset_stats}
\section{Experimental Setup}
\subsection{Dataset and Evaluation Metric}
Our experiments are conducted on a multilingual fact-checking dataset: \xfact~ \cite{gupta-srikumar-2021-x}. \xfact~ contains 31,189 claims in 25 languages collected from fact-checking websites via Google’s Fact Check Explorer. Each claim is annotated with one of the following seven labels: \textsc{True}, \textsc{Mostly-True}, \textsc{Partly-True}, \textsc{Mostly-False}, \textsc{False}, \textsc{Unverifiable}, and \textsc{Other}. \citet{gupta-srikumar-2021-x} split the data in a way that allows evaluation in various settings, as described in \Cref{tab:dataset_stats}. \textit{In-domain} and \textit{out-of-domain} test sets contain claims in the same languages as those in the training set, except that the claims in the \textit{out-of-domain} split are from different websites. Our main focus is the \textit{zero-shot} setup, where there is no overlap between the languages in the \textit{zero-shot} split and those in the training set. We use macro F1 as the evaluation metric, following \citet{gupta-srikumar-2021-x}.% Detailed statistics of \xfact~ can be found in \Cref{sec:dataset_stats}.
\input{tables/main_results}

\subsection{Baselines}
We compare the following competitive retrieval systems using different retrieval components but with the same multilingual reader.

\paragraph{MT + DPR} A common approach to cross-lingual tasks is translating inputs from target languages to source languages of richer resources so that stronger monolingual models can be utilized \cite{ahmad2020gate, asai-etal-2021-xor}. We translate all claims and all passages into English with the \textsc{Helsinki-NLP} neural machine translation models\footnote{\url{https://huggingface.co/Helsinki-NLP}} for its comprehensive language coverage and decent performance. For languages not covered by \textsc{Helsinki-NLP}, we use Google Translate instead. Then, we use DPR \cite{karpukhin-etal-2020-dense} to perform retrieval based on the translated claims and passages.

\paragraph{BM25} BM25 \cite{robertson2009probabilistic} has demonstrated advantages over dense vector representation approaches in monolingual retrieval tasks \cite{lee-etal-2019-latent}. Since BM25 only works in a monolingual setup, we create dummy empty passages for claims whose languages are not presented in the passage collection $P$. Our implementation is based on the Rank-BM25 package\footnote{\url{https://pypi.org/project/rank-bm25/}}.
\paragraph{mDPR} mDPR is a multilingual retriever based on DPR. It was trained on multilingual question answering datasets, as detailed in \Cref{sec:method}.

\paragraph{Google Search} As demonstrated in previous work \cite{augenstein-etal-2019-multifc, gupta-srikumar-2021-x}, snippets from Google Search results can serve as evidence for fact-checking. We directly take the snippets obtained by \citet{gupta-srikumar-2021-x} as inputs.

\subsection{Implementation Details}
When trained with X-ICT on the passage collection, the retriever is optimized using AdamW \cite{loshchilov2018decoupled} with a learning rate of 2e-5 over 30 epochs. When fine-tuning the multilingual reader, we set the learning rate to 5e-5 for parameters in mBERT and 1e-3 for all other parameters. The maximum input sequence length for \xict~ and fine-tuning on \xfact~ are set to 256 and 512, respectively. We use the pre-trained mBERT checkpoints on HuggingFace\footnote{\url{https://huggingface.co/bert-base-multilingual-cased}}.

%% file: tables/dataset_stats.tex
\begin{table}[t]
    \small
    \centering
    {
    \begin{tabular}{lcc}
        \toprule 
        Split & \# claims & \# languages \\
        \midrule 
        Train         & 19079 & 13 \\
        Development   & 2535  & 12 \\
        In-domain     & 3826  & 12 \\
        Out-of-domain & 2368 & 4 \\
        Zero-shot    & 3381 & 12 \\
        \bottomrule
    \end{tabular}
    }
    \vspace{-2mm}
    \caption{Dataset statistics of \xfact~.}
    \label{tab:dataset_stats}
    \vspace{-5mm}
\end{table}

%% file: tables/main_results.tex
\begin{table*}[t]
    \small
    \centering
    {
    \begin{tabular}{cccccc}
        \toprule
        
        & \textbf{Reader} & \textbf{Retrieval Method} & \textbf{Zero-shot F1 (\%)} & \textbf{In-domain F1 (\%)} \\

        \midrule
        
        \multirow{3}{*}{\shortstack{Prior\\ \cite{gupta-srikumar-2021-x}}} 
        & Majority & None   & 7.6 & 6.9\\
        &  mBERT & None & 16.7 & 39.4\\
        & mBERT & Google Search & 16.0 & 41.9\\
        \midrule 
        \multirow{6}{*}{Ours} & mBERT & None & 17.25 & 36.91  \\
                              & mBERT & Google Search & 16.02 & \textbf{42.61}   \\
                              & mBERT & MT+DPR & 15.01 & 35.29 \\
                              & mBERT & BM25 & 17.43 & 38.29\\
                              & mBERT & mDPR  & 17.60 & 36.79\\
                              & mBERT & \modelshort~ & $~~\textbf{19.83}^*$ & 40.53 \\

        \bottomrule
    \end{tabular}
    }
    \vspace{-2mm}
    \caption{Performance comparison in macro F1 (\%) of various models on the \textsc{X-Fact} test sets. \textit{None} retrieval method means not using any retrieval component, while \textit{Majority} means predicting the majority label, \textsc{False}, for all samples. When no retrieval method is used, the reader performs fact checking simply based on the claim template $T$ described in \Cref{sec:method}. Statistical significance over the second best models computed using the paired bootstrap procedure \cite{berg-kirkpatrick-etal-2012-empirical} is indicated with $^*$ ($p < .05$).}
    \label{tab:main}
    \vspace{-5mm}
\end{table*}

%% file: contents/06_results.tex
\section{Results}
\label{sec:results}
\subsection{Main results}
\Cref{tab:main} summarizes the fact-checking performance on the \xfact~ dataset. Our framework establishes a new state-of-the-art in zero-shot cross-lingual fact-checking, outperforming the previous best models by an absolute macro F1 of 2.23\%. The improvements demonstrate the effectiveness of our approach in retrieving relevant passages from multiple languages to assist in fact-checking. Furthermore, we found that the use of vanilla mDPR does not improve performance compared to the absence of any retrieval component. This can be explained by the domain discrepancy issue for mDPR, as it was trained on question-like queries instead of claim-like queries. Furthermore, although \modelshort~ is the most advantageous in the zero-shot setup, it trails behind Google Search in-domain setup. The significantly increased gap between in-domain F1 score and zero-shot F1 score for using Google Search suggests that the reader may exploit biases or patterns presented in Google Search's results that are not transferrable across languages. To validate this hypothesis, we analyzed the relationship between the snippets returned by Google Search and the ground-truth labels. We found that for claims of richer resources in the training set, Google Search is able to retrieve evidence that strongly indicates the veracity of the claims due to the abundance of fact-checking websites. For example, among Indonesian claims where Google Search results contain the string ``SALAH'' (WRONG), 50\% of them are \textsc{Partly True} and 45\% of them are \textsc{False}. Such patterns can also be found in the \textit{in-domain} split, but not in the \textit{zero-shot} split. This finding explains the increased gap between the performance on these two splits when using Google Search as the retrieval method. It also implies that our approach is more generalizable to lower-resource languages and more applicable when no fact-checking websites have debunked the input claims.

\subsection{Performance Analysis}
\label{subsec:performance_analysis}
\begin{figure}[t]
    \centering
    \includegraphics[width=0.9\linewidth]{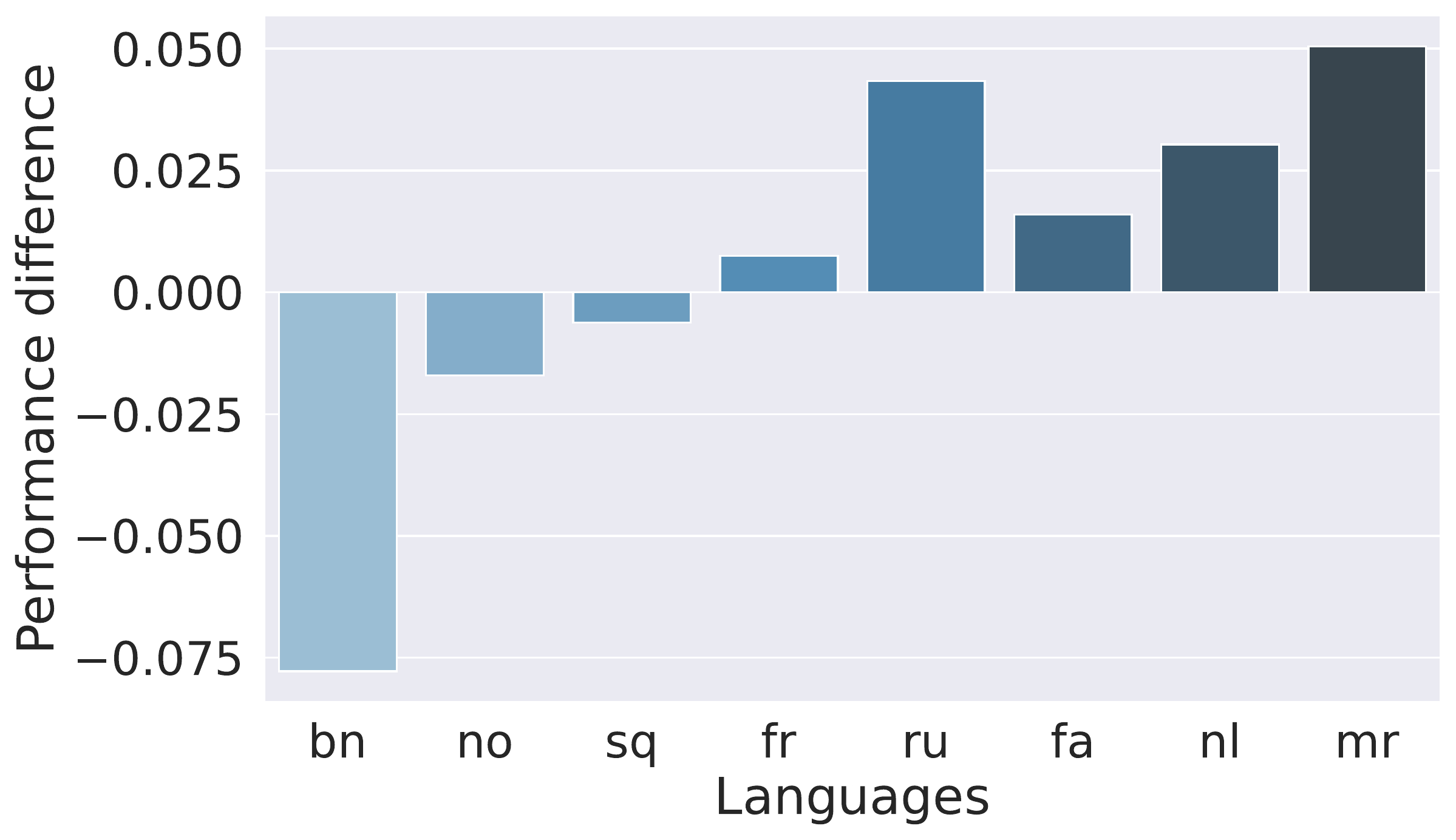}
    \vspace{-2mm}
    \caption{Performance difference in macro F1 when Indonesian passages are removed from the passage collection. On the x-axis, languages are sorted in ascending order by distance to Indonesian (i.e. Bengali is the closest, while Marathi is the furthest). We compute the distance between two languages based on word ordering, following \citet{ahmad-etal-2019-difficulties}.}
    \vspace{-5mm}
    \label{fig:ablate_id}
\end{figure}

\paragraph{Language Choice in Passage Collection}
Throughout our experiments, we found that the language of the retrieved passage can affect the fact-checking performance. We hypothesize that it may be more challenging for readers to reason through passages when the passage language is distant from the claim language. To verify such a hypothesis, we first compute the distance between each pair of languages based on word ordering, following \citet{ahmad-etal-2019-difficulties}. Then, we remove passages of a particular language from the passage collection. With this subset of selected passage collection, we use \modelshort~ to retrieve relevant passages for fact-checking and train another fact-checker based on newly retrieved passages. Finally, we compare the performance difference for different languages between this system and the model discussed in \Cref{sec:method} in the \textit{zero-shot} setup. \Cref{fig:ablate_id} shows the results for each language when Indonesian passages are removed from the passage collection. We observe that the performance drop and the distance between each language and Indonesian are highly correlated. In fact, we see that removing Indonesian is beneficial for some distant languages such as Dutch and Marathi. 

\begin{figure}[t]
    \centering
    \includegraphics[width=0.9\linewidth]{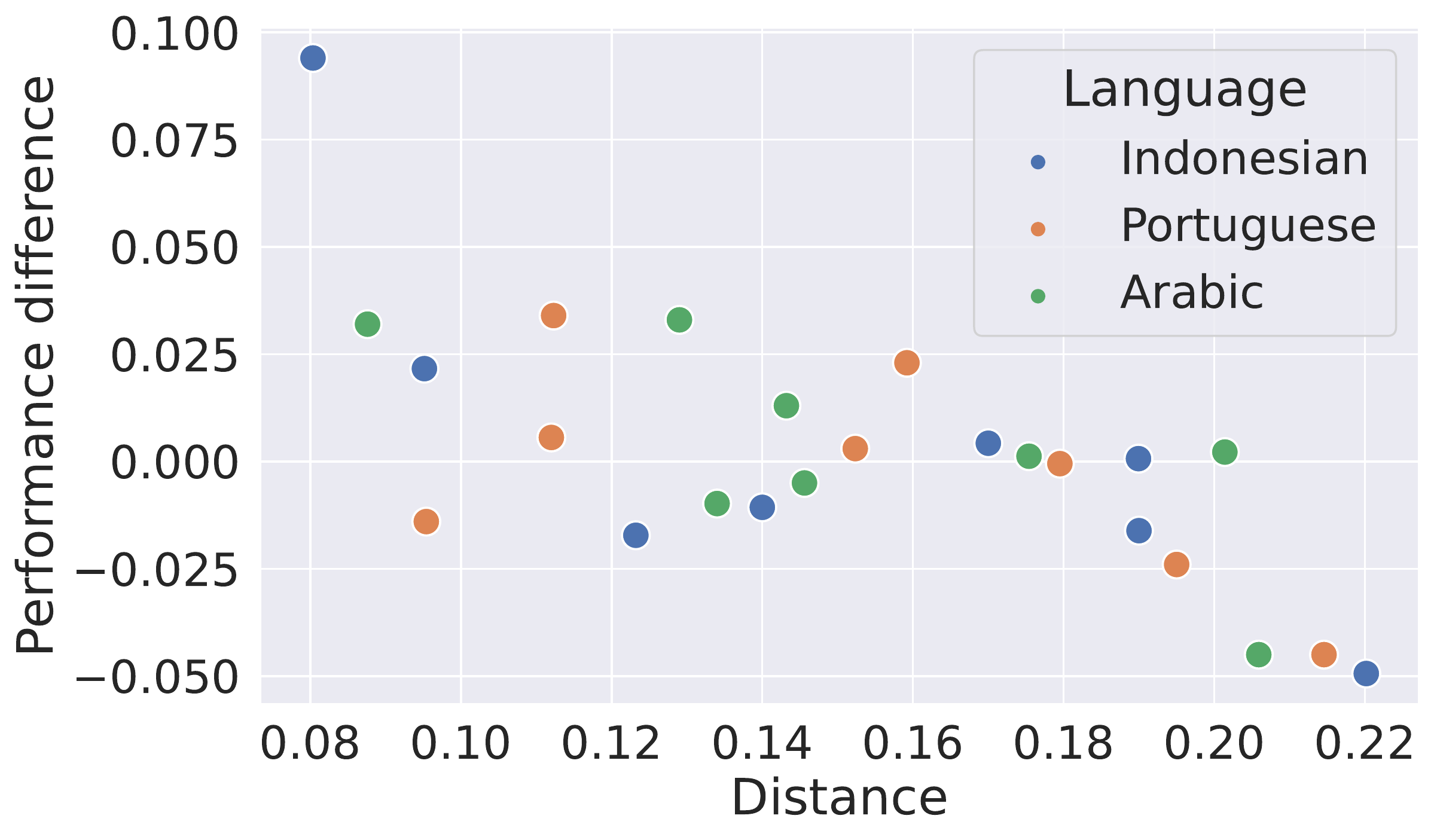}
    \vspace{-2mm}
    \caption{Performance difference in macro F1 when the retrieved passages are translated to Indonesian, Portuguese, and Arabic, respectively. Overall, the performance difference has a negative correlation with the distance between the passage's language and the claim's language. }
    \vspace{-5mm}
    \label{fig:ablate_three_langs}
\end{figure}

Given these results, the following question arises: is this phenomenon caused by the poor capability of our multilingual reader to reason with passages and claims in distant languages, or by the higher information overlap between Indonesian passages and claims in closer languages? To better understand the results, we translate the retrieved passages into three languages of distinct language families: Indonesian, Portuguese, and Arabic. With the three sets of translated passages, we train three fact-checkers. Then, we compute the difference between the original performance and the performance achieved using the translated passages. In \Cref{fig:ablate_three_langs}, we observe that the performance difference is negatively correlated with the distance between the languages of the claim and the passage. The correlation coefficient is -0.490 per Kendall's Tau \cite{kendall1938new}. This confirms that passages in distant languages are indeed harder for our multilingual reader to reason with. The trend is consistent with the findings of \citet{asai2021cora}.

\paragraph{Impact of the Amount of Training Data}

\begin{figure}[t]
    \centering
    \includegraphics[width=0.9\linewidth]{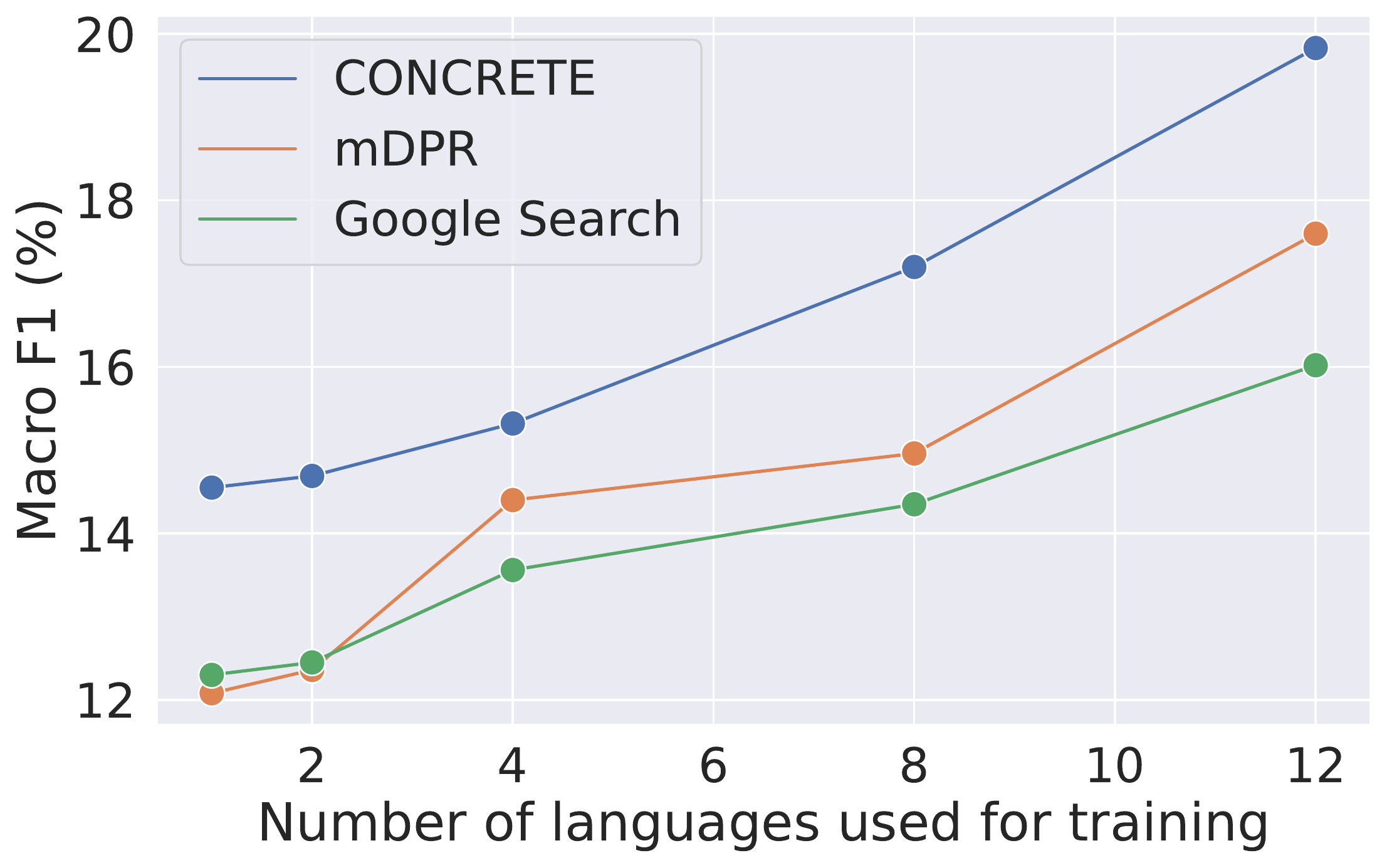}
    \vspace{-2mm}
    \caption{Zero-shot cross-lingual performance on \xfact~ with regard to various numbers of languages used for training.}
    % \vspace{-5mm}
    \label{fig:data_efficiency}
\end{figure}

To test the data efficiency of our approach, we compared \modelshort~ with mDPR and Google Search in the \textit{zero-shot} setup using different numbers of languages for training. As shown in \Cref{fig:data_efficiency}, \modelshort~ consistently outperforms the other two methods across all settings. This indicates the strength of \modelshort~ in aiding fact-checking in low- and high-resource scenarios.

\begin{figure}[t]
    \centering
    \includegraphics[width=0.9\linewidth]{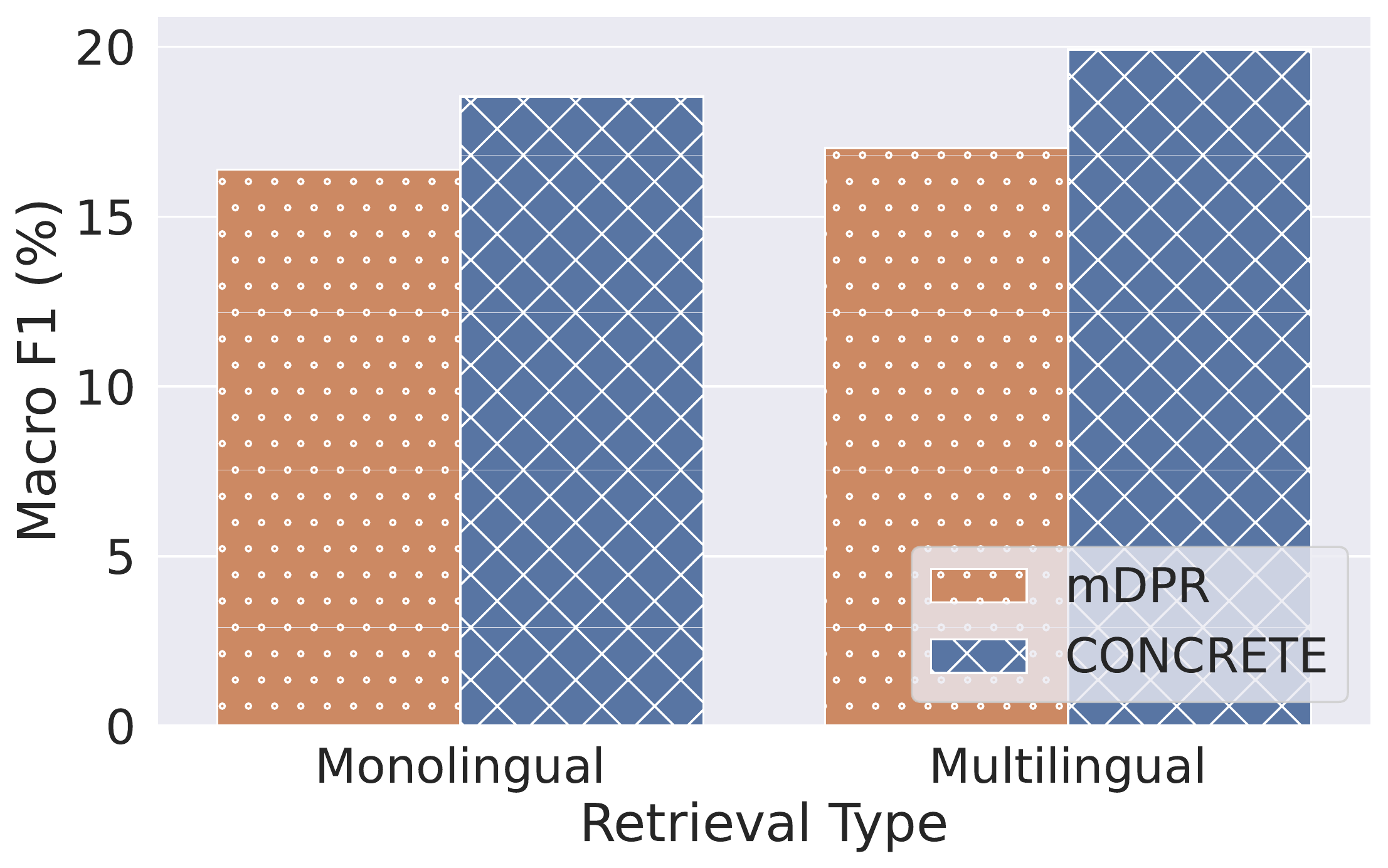}
    \vspace{-2mm}
    \caption{Performance comparison between using monolingual and multilingual retrieval on cross-lingual fact-checking. For both mDPR and \modelshort~, retrieving passages in multiple languages improves the overall performance.}
    % \vspace{-5mm}
    \label{fig:monolingual_vs_multilingual}
\end{figure}

\paragraph{Impact of Retrieving from Multiple Languages}
We conducted a case study on cross-lingual retrieval versus monolingual retrieval to understand whether retrieving passages from multiple languages actually helps the performance. In particular, we train and evaluate the models on samples whose languages are in the passage collection\footnote{We train the models on Portuguese, Indonesian, and Arabic, and evaluate them on French, Persian, and Russian.}. For the monolingual retrieval setting, the model is restricted to retrieving passage in the same language as the input claim, while the cross-lingual setting does not have such a restriction. We found that for both mDPR and \modelshort~, cross-lingual retrieval setting outperforms their monolingual retrieval counterparts, as shown in \Cref{fig:monolingual_vs_multilingual}. This finding confirms that retrieving evidence in multiple languages helps cross-lingual transfer for fact-checking.

\subsection{Qualitative Analysis}
\label{subsec:qualitative_analysis}
The following qualitative analysis provides an intuition for our model's advantage in cross-lingual fact-checking.

\begin{figure*}[t]
    \centering
    \includegraphics[width=0.9\linewidth]{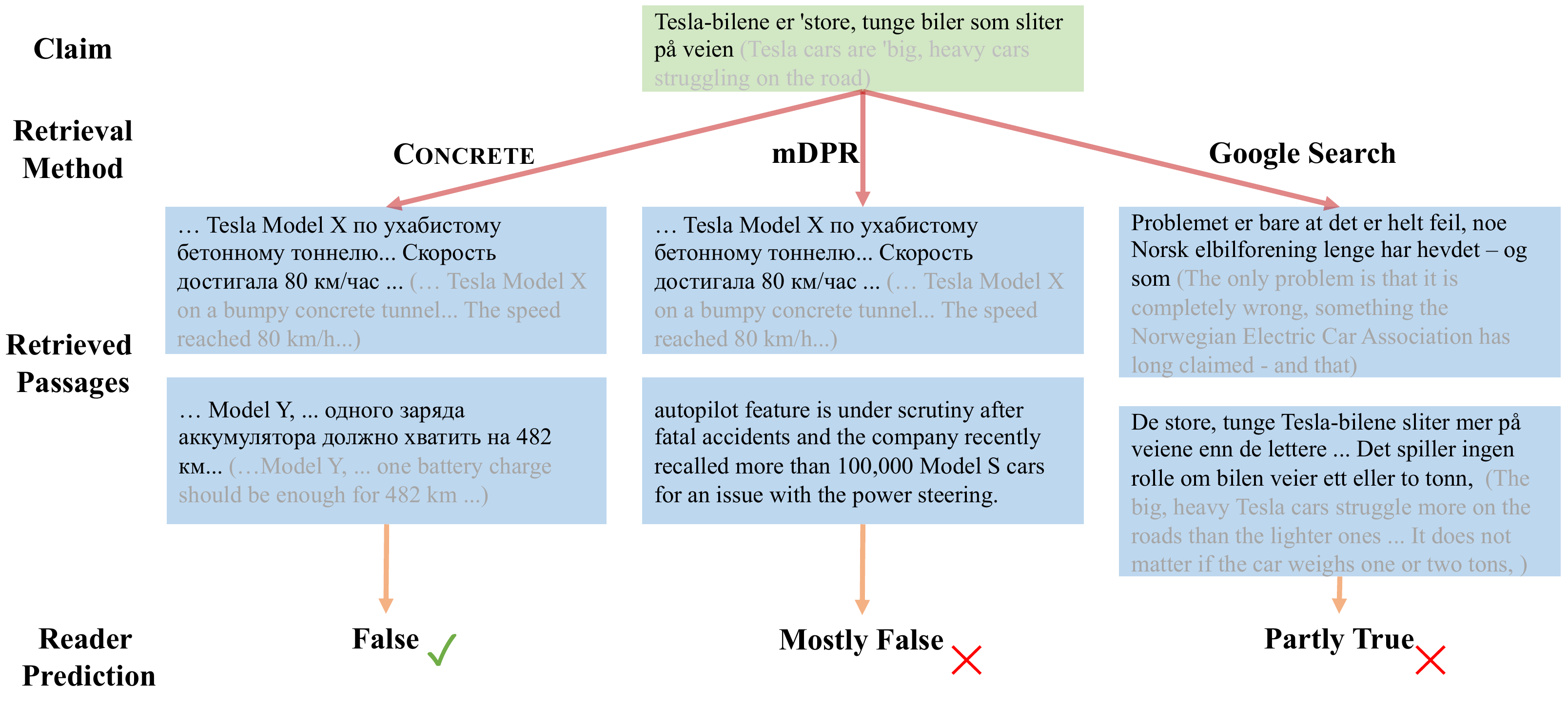}
    \vspace{-2mm}
    \caption{An example showing how relevant trustworthy passages retrieved by \modelshort~ lead to correct predictions. \modelshort~ retrieves two passages that prove Tesla cars are not slow or struggling on the road. Since the passages are from trustworthy sources, our reader can correctly predict the claim as \textsc{False}. On the other hand, the second passage that mDPR retrieves is slightly related to the claim (Tesla) but not directly relevant to the topic (the poor performance of Tesla cars). For Google Search's results, these two passages contradict each other as one is from a fact-checking website and the other is from an unreliable source. Hence, the model cannot predict correctly based on the retrieved passages from mDPR or Google Search.}
    % \vspace{-5mm}
    \label{fig:qualitative_analysis}
\end{figure*}

\paragraph{Impact of \xict~}
To validate the effectiveness of \xict~ in retrieving claims that are more relevant to the topic of the claim, we compared 50 predictions between models using \modelshort~ and mDPR as retriever in the split \textit{zero-shot}. The results show that 23 errors made by mDPR are corrected by \modelshort~, while only 6 new errors are introduced. We found that mDPR can often retrieve passages that are relevant to a part of a claim, but the topic of the retrieved passages may not align well with that of the claim, likely due to the domain mismatch issue discussed in \Cref{sec:method}. An example is shown in \Cref{fig:qualitative_analysis}. This reflects that \xict~ is able to improve the retrieval quality even though only pseudo feedback is used for training.
\paragraph{Importance of Passages' Trustworthiness} Comparing the predictions and the retrieved passages between \modelshort~ and Google Search on the \textit{zero-shot} split, we observe that our approach is better at identifying \textsc{False} and \textsc{Mostly False} claims. As demonstrated in \Cref{fig:qualitative_analysis}, when a claim is \textsc{False} or \textsc{Mostly False}, the snippets returned by Google Search are often contradicted with each other, which usually leads to incorrect predictions. The inconsistency in the snippets from Google Search is caused by the fact that Google Search retrieves information from the entire Web without considering the trustworthiness of the source. On the contrary, our approach offers the flexibility to include only trustworthy information in the passage collection. In the \textit{zero-shot} split, we found that this property of Google Search leads to 44 and 67 more errors in identifying \textsc{False} and \textsc{Mostly False} claims, respectively.

\subsection{Remaining Challenges}

\begin{figure}[h]
    \centering
    \includegraphics[width=0.9\linewidth]{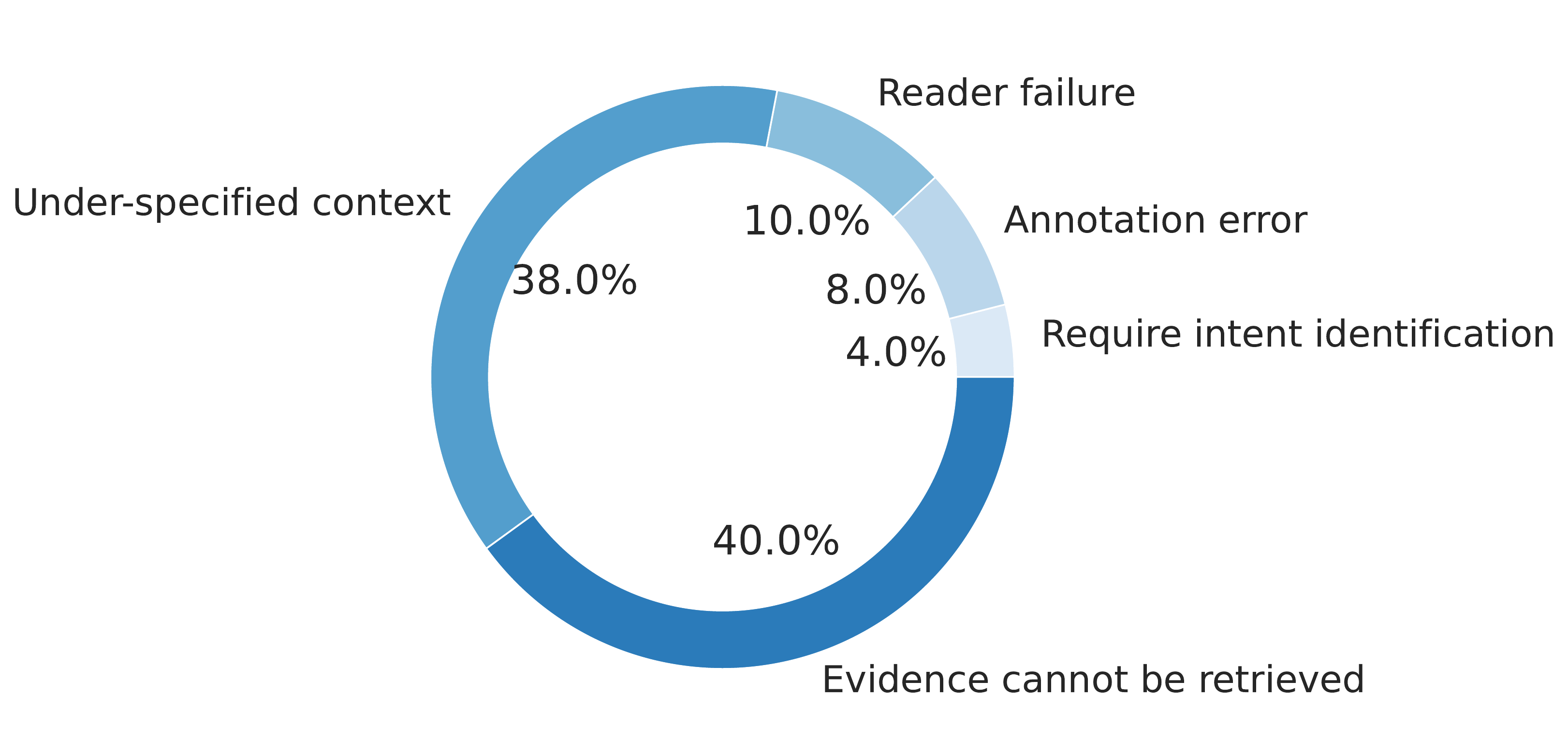}
    \vspace{-2mm}
    \caption{Distribution of the remaining errors.}
    % \vspace{-5mm}
    \label{fig:remaining_challenges}
\end{figure}

To identify the remaining challenges, we compare 50 errors made by our model with ground-truth labels and analyze the sources of errors, as illustrated in \Cref{fig:remaining_challenges}. The following paragraphs will discuss these categories with examples.

\paragraph{Evidence cannot be retrieved.} The most common error is caused by the absence of supporting or refuting evidence in the passage collection. For the majority of such errors, the claims are country specific. For example, 
 \begin{quoting}
     Mann med tre koner får tre leiligheter i Sverige.
     (Husband with three wives gets three apartments in Sweden.)    
\end{quoting}
We can address this issue by adding trustworthy news articles from more countries into the passage collection. 

\paragraph{Under-specified context.} Another major source of errors is the underspecification of the input claim. An example claim is:
\begin{quote}
    ``Nuk besoj që janë të informuar as partnerët tanë, SHBA dhe NATO, sepse do të isha i informuar edhe unë.'' (I do not believe that our partners, the US and NATO, are informed either, because I would be informed as well.)    
\end{quote}
 In this claim, it does not specify who is ``I'' and what the US and NATO are not informed. Therefore, the given information is too little to determine the veracity of the claim. This problem could potentially  be solved by mining the original context from the Internet. 

\paragraph{Require intent identification.}
Some of the claims are correct but contain misleading information. For instance,

\begin{quoting}
    \foreignlanguage{russian}{``Писатель Дин Кунц предсказал появление коронавируса в своей книге в 1981 году и называл его Ухань-400.''} 
    (Writer Dean Koontz predicted the emergence of the coronavirus in his book in 1981 and called it Wuhan-400.)
\end{quoting}
This claim is correct, but the claimer attempts to mislead the audience by linking the coronavirus in the book with COVID-19, which is false. To correctly predict this claim, the model should be capable of identifying the intent behind the claims.

\paragraph{Reader failure.} In some cases, the reader fails to predict the correct veracity even though the supporting evidence is successfully retrieved. This is mainly due to the long distances between passages and claims, as discussed in \Cref{subsec:performance_analysis}.

\paragraph{Annotation error.} The dataset was created by matching the rating of each claim on fact-checking websites with its label definition. Annotation errors could be caused by (1) the misalignment between the rating and the label definition, and (2) the inaccurate ratings listed on the fact-checking websites. We found that each case accounts for about half of the annotation errors.

%% file: contents/02_related_work.tex
\section{Related Work}
\subsection{Fact-checking}
Previous fact-checking approaches can be roughly divided into two categories based on task formulations. The first type of formulation assumes that evidence candidates are given, such as the \textsc{FEVER} dataset \cite{thorne-etal-2018-fever} and the \textsc{SciFact} dataset \cite{wadden-etal-2020-fact}. Previous approaches for this category of fact-checking tasks often involve a retrieval module to retrieve relevant evidence from the given candidate pool followed by a reasoning component that determines the compatibility between a piece of evidence and the input claim \cite{yin-roth-2018-twowingos, pradeep-etal-2021-scientific}. The second category is the \textit{open-retrieval} setting\footnote{We borrow the term \textit{open-retrieval} from the field of question answering.}, where evidence candidates are not provided, such as the \textsc{Liar} dataset \cite{wang-2017-liar} and the \textsc{X-Fact} dataset \cite{gupta-srikumar-2021-x}. For this task formulation, one of the main challenges is where and how to retrieve evidence. Some work determines the veracity of a claim based solely on the claim itself and the information learned by language models during the pre-training stage \cite{lee-etal-2021-towards}. However, such an approach is tied to the period of time in which the pre-training data is collected and does not generalize well to new claims. Other studies devise hand-crafted linguistic features as input to the fact-checking models \cite{mihalcea-strapparava-2009-lie, choudhary-and-arora-2021}. However, these approaches are language-specific, whereas our approach is language-agnostic since it consists of a cross-lingual retriever and a multilingual language model. \citet{gupta-srikumar-2021-x} use a similar approach, which retrieves relevant snippets using Google Search instead of a cross-lingual retriever. Our experimental results show that the proposed cross-lingual retriever is more effective than Google Search in the \textit{zero-shot} setting due to the fact that Google Search ignores the trustworthiness of the retrieved information and that downstream models tend to exploit the biased patterns in Google Search results that are not transferrable across languages, as shown in \Cref{sec:results}.

\subsection{Cross-lingual Retrieval}
Early attempts on cross-lingual retrieval adopt a pipeline consisting of a statistical machine translation system and a monolingual retrieval model \cite{hiemstraj-1999-disambiguation, ture-2013-flat}. These methods do not perform well due to the poor performance of statistical machine translation systems. Later work addresses this issue with bilingual embeddings \cite{Vulic2015MonolingualAC, Litschko2018UnsupervisedCI}. More recently, large pre-trained multilingual language models demonstrate significant advantages in constructing multilingual representations \cite{jiang-etal-2020-cross}. \citet{Yu2021CrosslingualLM} pre-train a cross-lingual language model tailored for the retrieval tasks. Yet, the computation complexity is relatively high due to the cross-encoder architecture of the model. Namely, it takes a pair of query and evidence as inputs, instead of encoding the query and evidence independently. The mDPR model presented in \cite{asai2021cora} is the most favorable for our task due to its high efficiency and performance. However, mDPR was trained on datasets where queries are questions instead of claims. Therefore, domain adaptation is needed for mDPR to be applied to our task. \modelshort~ extends mDPR by adapting it to fact-checking using a self-supervised cross-lingual retrieval algorithm to mitigate the domain discrepancy problem while maintaining high efficiency.

%% file: contents/07_conclusion.tex
\section{Conclusions and Future Work}
We have proposed \modelshort~, a claim-oriented cross-lingual retriever that retrieves trustworthy passages from a multilingual passage collection. To overcome the lack of IR training data with claim-like queries, we present the Cross-lingual Inverse Cloze Task (\xict~) that leverages pseudo feedback to train the retriever. Experimental results on \xfact~ showed that our approach outperforms all previous systems in the \textit{zero-shot} cross-lingual setting. For future work, we plan to investigate the adaptive selection mechanism for passages based on distances and develop more robust readers for reasoning through passages of longer distances via representation learning.

%% file: contents/08_ethics.tex
\section{Ethical Considerations}
Although our framework has significant advantages over the previous state of the art, the proposed model is still far from being a reliable cross-lingual fact checker given its great potential for improvement. If such a system is deployed for public use, the general public could lose trust in automatic fact-checking systems, and the situation of infodemic can exacerbate. Therefore, at the current stage, our system should be served as an assistant for human fact-checkers to validate the veracity of claims instead of directly applying for public use, especially in the \textit{zero-shot} setting. With our framework, the efficiency of manual fact-checking can be significantly improved thanks to its ability to retrieve relevant information across multiple languages and produce a reasonably good preliminary judgement on the veracity of the input claim.

In addition, we also acknowledge that the use of large multilingual language models pre-trained on the Web could lead to biased outputs. Fortunately, after a close inspection into the \xfact~ dataset, we do not find such biased patterns in it. This means that fine-tuning our proposed framework on \xfact~ should alleviate the problem of biased predictions.